\documentclass{article} 
\usepackage[preprint]{colm2025_conference}

\usepackage{microtype}
\usepackage{hyperref}
\usepackage{url}
\usepackage{booktabs}
\usepackage{amsmath}
\usepackage{enumitem}       
\usepackage{soul}

\usepackage{lineno}

\definecolor{darkblue}{rgb}{0, 0, 0.5}
\hypersetup{colorlinks=true, citecolor=darkblue, linkcolor=darkblue, urlcolor=darkblue}

\newcommand{\rev}[1]{\textcolor{black}{#1}}

\newcommand{\uarr}{$\uparrow$}
\newcommand{\darr}{$\downarrow$}

\newcommand{\titleFullNameFirst}{Call of Duty\textregistered: Modern Warfare\textregistered III}
\newcommand{\titleFullName}{Call of Duty: Modern Warfare III}
\newcommand{\titleShort}{MWIII}

\definecolor{lightBlue}{rgb}{0.78, 0.85, 1.0}
\definecolor{lightOrange}{rgb}{0.88, 0.95, 1.0}
\definecolor{lightRed}{rgb}{1.0, 0.85, 0.85}
\usepackage{tcolorbox}

\newtcbox{\bluebox}{on line, box align=base, colback=lightBlue,colframe=white,size=fbox,arc=3pt, before upper=\strut, top=-2pt, bottom=-4pt, left=-2pt, right=-2pt, boxrule=0pt}

\newtcbox{\orangebox}{on line, box align=base, colback=lightOrange,colframe=white,size=fbox,arc=3pt, before upper=\strut, top=-2pt, bottom=-4pt, left=-2pt, right=-2pt, boxrule=0pt}

\newtcbox{\redbox}{on line, box align=base, colback=lightRed,colframe=white,size=fbox,arc=3pt, before upper=\strut, top=-2pt, bottom=-4pt, left=-2pt, right=-2pt, boxrule=0pt}

\newcommand{\dsetdota}{\textit{DOTA 2}}
\newcommand{\dsetcod}{\rev{\textit{\titleShort}}}

\newcommand{\base}{distilRoBERTa-base} %
\newcommand{\DAP}{+DAP} %
\newcommand{\sep}{+sep tokens} %
\newcommand{\sender}{+metadata sep tokens} %

\graphicspath{ {./images/} }

\title{Context-Aware Toxicity Detection in Multiplayer Games: Integrating Domain-Adaptive Pretraining and Match Metadata}

\author{Adrien Schurger-Foy\textsuperscript{1}, Rafal Kocielnik\textsuperscript{2}, Caglar Gulcehre\textsuperscript{1\textdagger}, R. Michael Alvarez\textsuperscript{2\textdagger}  \\
\textsuperscript{\textdagger}Last author
\textsuperscript{1}EPFL
\textsuperscript{2}Caltech
}

\begin{document}

\ifcolmsubmission
\linenumbers
\fi

\maketitle

\begin{abstract}
The detrimental effects of toxicity in competitive online video games are widely acknowledged, prompting publishers to monitor player chat conversations. This is challenging due to the context-dependent nature of toxicity, often spread across multiple messages or informed by non-textual interactions. 
Traditional toxicity detectors focus on isolated messages, missing the broader context needed for accurate moderation. This is especially problematic in video games, where interactions involve specialized slang, abbreviations, and typos, making it difficult for standard models to detect toxicity, especially given its rarity. 
We adapted RoBERTa LLM to support moderation tailored to video games, integrating both textual and non-textual context. By enhancing pretrained embeddings with metadata and addressing the unique slang and language quirks through domain adaptive pretraining, our method better captures the nuances of player interactions. 
Using two gaming datasets - from Defense of the Ancients 2 (\dsetdota{}) and \titleFullNameFirst{} (\dsetcod{}) we demonstrate which sources of context (metadata, prior interactions...) are most useful, how to best leverage them to boost performance, and the conditions conducive to doing so. 
This work underscores the importance of context-aware and domain-specific approaches for proactive moderation.
\end{abstract}

\section{Introduction}

Online platforms that require or benefit from moderation almost always involve text conversations. In these cases moderators would like to check the text content before it is posted, allowing them to act proactively (e.g. filtering) before any negative impact occurs. However it is not always easy to assess a message or comment without understanding the context in which it was posted. This is especially true in multiplayer video games where much of the interaction between players exists outside of a given text conversation or message.

Swift moderation is known to be more effective \citep{pratt2017empirical, abramowitz1990effectiveness,srinivasan2019content}, and can de-escalate toxic interactions that could otherwise spread among other players \citep{morrier2024uncovering} and possibly outside a given platform \citep{OnlineHa73:online}. Online platforms have an incentive to apply effective moderation not only to maintain a positive reputation, but also because toxicity can reduce engagement \citep{kocielnik2024challenges,morrier2024uncoveringB}. The context of text content is essential for catching implicit toxicity, a fact users can exploit to evade moderation \citep{li2022adversarial, ma2023users}.

\subsection{Challenges}
Most toxicity classifiers only consider a single message or post \citep{sheth2022defining}, but the datasets used in this work contain many cases where toxicity is spread over multiple messages, while individual messages are not toxic in a vacuum. In these cases, the toxicity is undetectable (by annotator or classifier) if only one message is considered. Conversely, if the entire chat history is used as the input document, a typical classifier would lack specificity (it can tell if the match is toxic but not which messages). Token and span detection approaches have specificity, but would flag past messages and so do not support proactive moderation \citep{chhablani2021nlrg}.

Since toxicity lacks an objective definition, platforms must determine moderation thresholds based on user demographics and expectations \citep{beres2021don}. In \rev{online competitive action} games, censoring \rev{all strongly worded} language entirely would leave little room for communication \citep{kwak2015exploring}. While excessive moderation (e.g., over-censoring) can reduce apparent toxicity \citep{li2024online}, it may also provoke adversarial adaptation, where players modify their language to evade detection while maintaining toxic intent \citep{warner2024critical}. Moreover, such restrictive policies can lead to increased frustration, which may manifest in other forms of negative behavior \citep{kosa2022need}.

\begin{figure}[t]
\begin{center}
\includegraphics[width=.75\textwidth]{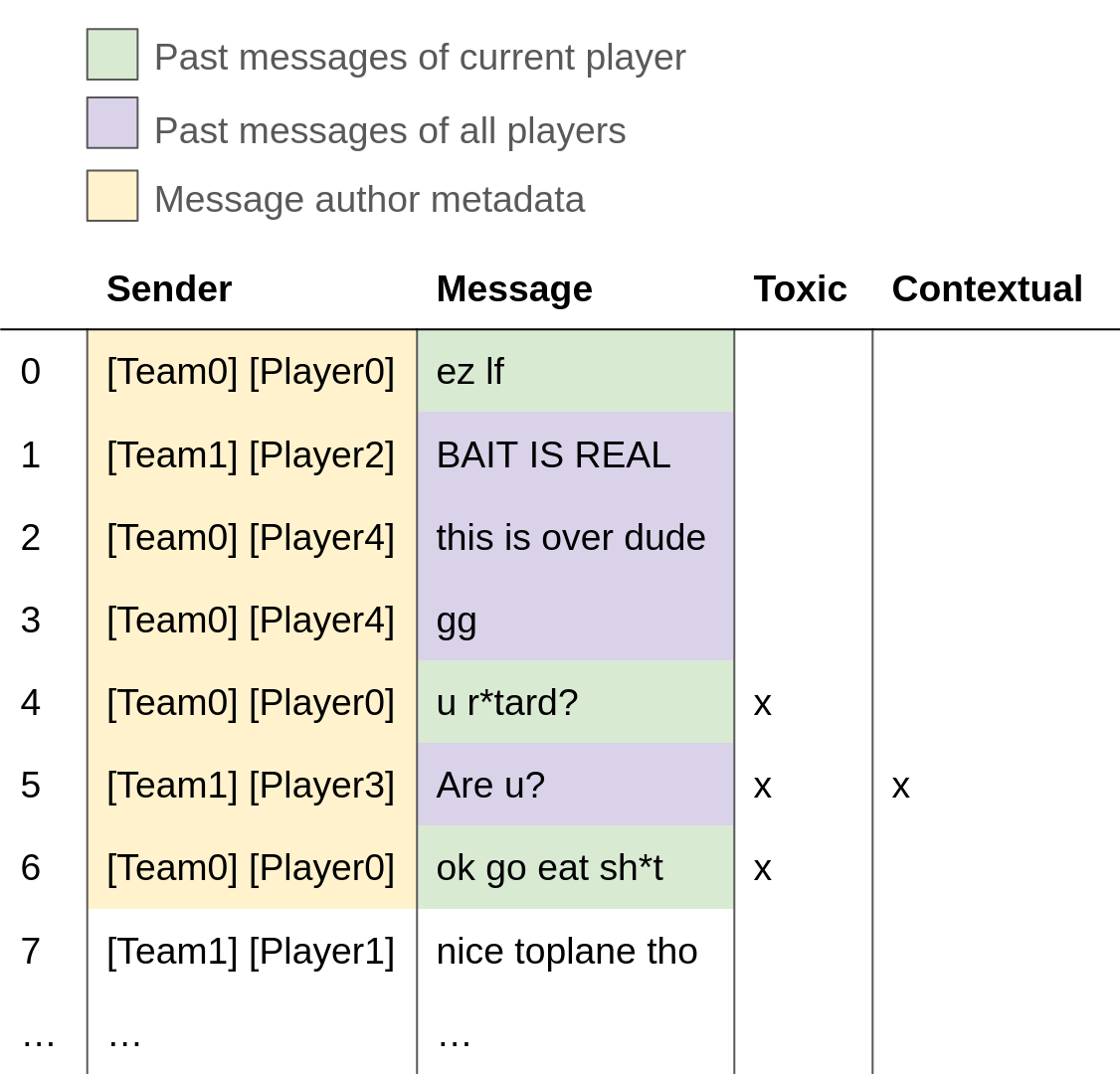}
\end{center}
\caption[]{Context levels within a match. Message 6 is evaluated for toxicity, 0 is the start of the match. In addition to the 'toxic' label, the annotator records whether or not the message is toxic only because of previous messages ('contextual' label).
}
\label{context_levels}
\vspace{0pt}
\end{figure}

Even with a suitable definition of toxicity, moderators or annotators may have different access to information leading to additional disagreement or ambiguity in what is considered toxic \citep{beres2021don}. Furthermore, it can be prohibitively costly to monitor or log \citep{li2020qualitative} the entire context surrounding a message or post, meaning classifiers and moderators might lack useful information. 

The process of annotating data to train text classifiers is also expensive, and in the case of toxicity detection, there is usually a large class imbalance (toxicity is rare) \citep{kocielnik2024challenges}. These two points result in fewer positive training samples, hindering machine learning. 

Finally, video games in particular often have their own specific vocabulary including character and item names, as well as slang, abbreviations, and typos \citep{zisa2016creation}. This results in data that is likely out of distribution for general pretrained foundation models, thus building classifiers on top of them may not be as successful as in typical situations, requiring many more labeled samples \citep{petrov2024language}.

\subsection{Approach}
The user written text a moderator would ideally like to review is a discrete segment of a conversation (such as a post, message, or comment), that can be acted on as it is being submitted. As such, rather than searching for a toxic span or set of toxic tokens in a conversation, we formulate the task in the following way: given the context (previous conversation, etc.), is the message about to be posted toxic? 

We employ Domain Adaptive Pretraining (DAP) \citep{gururangan2020don} using Masked Language Modeling (MLM) \citep{salazar2020masked} to capture the specific language used in gaming communities, including slang, abbreviations, and other domain-specific expressions. We integrate contextual information as additional special tokens in both MLM pretraining and the downstream task.  Through an ablation study, we demonstrate which combinations of metadata injection, pretraining, and past text interactions are most effective in boosting performance.

\subsection{Findings}

\begin{figure}[t]
\begin{center}
\includegraphics[width=\textwidth]{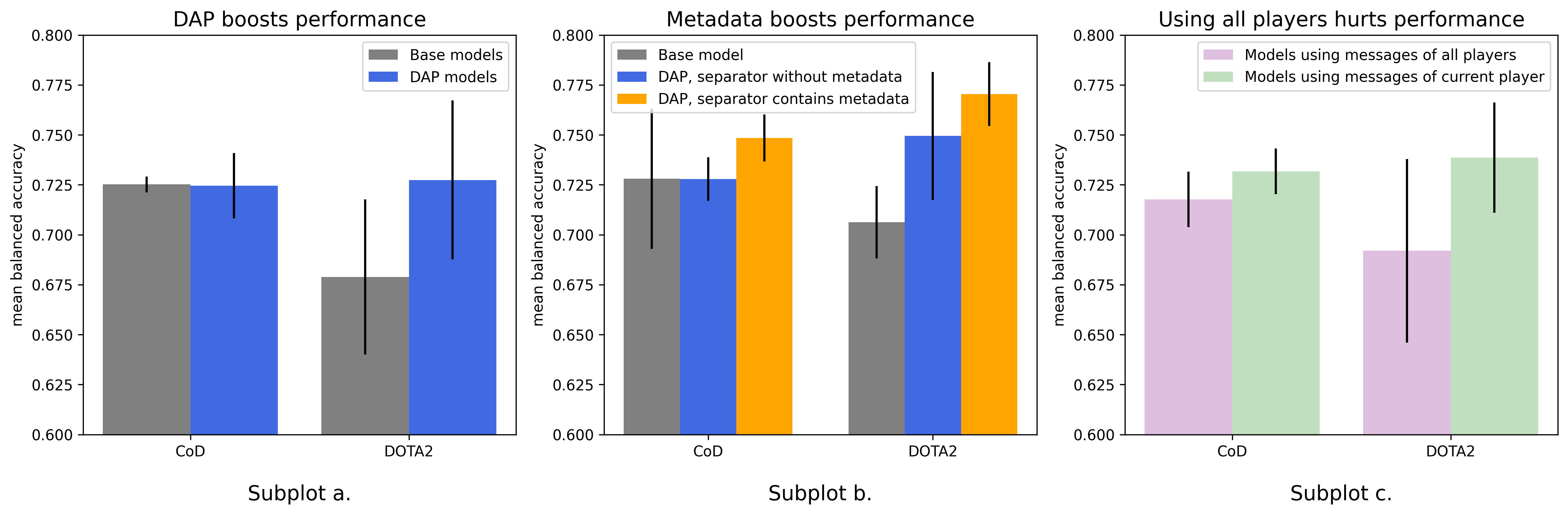}
\end{center}
\caption[]{Differing performance boost of DAP between datasets (subplot a.), performance boost from using metadata only in pretraining (subplot b.), more context is not always better (subplot c.). Subplots a. and c. are averaged across experiments, the error bar being standard deviation between experiments. Subplot b. is averaged across repeated runs of single experiments, the error bar being standard deviation between runs.}\label{takeaways}
\end{figure}

\begin{itemize}[leftmargin=*, itemsep=0.0mm, topsep=5pt]
    \item \textbf{Toxicity is highly contextual.}As much as 67\% of all toxic messages across both datasets are only toxic because of previous messages (e.g. message 5 in figure \ref{context_levels}). In other words, most instances of toxicity are not inherently toxic on their own but require contextual information to be accurately identified as such.
    
    \item \textbf{The efficacy of Domain Adaptive Pretraining (DAP) is situational.} The methods involving DAP were more effective on the \dsetdota{} dataset (figure \ref{takeaways} subplot a.). The conversations occurring in the \dsetcod{} dataset consist of a mix of voice and text chat, but only text chat was available to us. DAP did not work as well in this case, as too much of the conversation is unobserved.

    \item \textbf{Metadata is useful in both pretraining and finetuning.} We found that including metadata in pretraining improved performance, even when metadata information is not explicitly provided to the downstream classifier (figure \ref{takeaways} subplot c.). We conclude that metadata allows for higher quality pretrained embeddings.

    \item \textbf{More context is not always better.} In our case with few labeled training samples (100 matches per dataset), the classifier is not able to benefit from the messages of other players (figure \ref{takeaways} subplot c.). Using all players can be seen as multiplying the number \rev{of} input features, while the number of training samples remains small, leading to overfitting.

    \item \textbf{Long-term player behavior is predictive of current toxicity.} We achieve up to 74\% balanced accuracy in predicting a player's current toxicity using only their messages from other matches around the same time. This suggests that the past toxic behavior of a player is predictive of continued toxicity.
\end{itemize}

\begin{table}[t]
\begin{center}
\begin{tabular}{ p{.2\linewidth} p{.7\linewidth} }
    \toprule
    \textbf{Label} & \textbf{Label description} \\
    \midrule
    Toxic & Given the context, the message is hateful, disrespectful, or otherwise likely to discourage other players from engaging with this player. Taunting (calling a player bad at the game) is considered toxic if repeated constantly.\\
    \midrule
    Not toxic & Given the context, the message is unlikely to discourage other players from engaging with this player. Profanities not implicitly or explicitly directed at another player fall in this category.\\
    \hline
    \midrule
    Context-independent & The message would be considered toxic even without the previous messages. \\
    \midrule
    Context-dependent & The message is only toxic because of the previous messages, and would not be toxic alone. \\
    \bottomrule
\end{tabular}
\end{center}
\caption{We binarize and adapt Google perspective API's definition of toxicity (Appendix \ref{perp_tox_def}) to be more suited to our datasets (top two rows). We also flag where context is used for annotation (bottom two rows).}
\label{my_tox_def}
\end{table}

\section{Related work}
While keyword-based methods are still widely used in many practical applications \citep{jhaver2021evaluating, jhaver2022designing}, pretrained transformer-based toxicity detectors, such as Toxic-BERT \citep{Detoxify}, are readily available. Toxic-BERT doesn't have the limitations of keyword based methods (word sense ambiguity, implicit toxicity...), but it is designed to assess single documents, not to handle conversations.

Large general language models such as GPT-4 \citep{achiam2023gpt} should have the capability to overcome the challenges of conversational dialogue, but there are many issues that make their use impractical in this situation, cost being the most obvious. There are few entities that run such models, meaning platforms would likely face privacy issues by sending their users data externally, and have no control over changes to the model. Furthermore, prompt engineering to produce the desired definition of toxicity is challenging, as these models have deeply ingrained biases regarding what is toxic \citep{jan2025multitask}. Finally, there is an artificial barrier to using large general language models for toxicity detection: guardrails are usually built into these models that prevent them from processing extremely toxic content at all \citep{aldahoul2024advancing}. 

There are two works closest to ours in this space. \citet{weld2021conda} presents a dataset for which annotators consider previous conversation history, but don't develop any specific method for using this context in real-time classification. Furthermore, they offer no insight into how the annotator's consideration of previous messages affects ground truth labels. ToxBuster from \cite{yang2023toxbuster} is a transformer based classifier that is designed to use context (conversation history, message author, and recipients). We present new ways of using this context that enhance the quality of pretrained embeddings rather than the downstream classifier, and explore other sources of context derived from players' past interactions.

\section{Methods and data}
\subsection{Toxicity definition}
We adapted Google perspective API's definition of toxicity (table \ref{perp_tox_def}), as it is centered around engagement, an important objective of moderation. We binarize the definition, the two labels being ``toxic'' and ``non-toxic'' (table \ref{my_tox_def}). We also record whether or not a message is toxic in and of itself or if it is toxic only because of previous messages, although this only serves to provide insight into the contextual nature of toxicity and is not involved in classification.

We used two datasets, one from the game Defence of the Ancients 2 (\dsetdota{}) publicly available on Kaggle, and a proprietary dataset from \titleFullName{} (\dsetcod{}) provided by Activision. The datasets are primarily in English but also contain other languages, and contain only messages that all players in the match can see (no direct messages).

\begin{table}[ht]
\begin{tabular}{llrrr}

 &dataset & matches & messages & words \\
\midrule
unlabeled &\href{https://www.kaggle.com/datasets/romovpa/gosuai-dota-2-game-chats}{DOTA 2 - chat only\textsuperscript{1}} & 978,822 & 21,659,448 & 49,397,435 \\
 &\href{https://www.kaggle.com/datasets/devinanzelmo/dota-2-matches}{Dota 2 chat + game data\textsuperscript{2}}& 49,772 & 1,439,488 & 3,465,154 \\
  &\dsetcod& 1,470,796& 28,891,749&71,727,165\\ 
\midrule
labeled & \dsetdota& 300& 8,968&21,149\\
 & \dsetcod& 200& 3,677&9,034\\
\end{tabular}
\caption[]{
Datasets used for finetuning and domain-adaptive pretraining (DAP). \rev{Our datasets are subsets of the full data and do not represent the entirety of the player volume or experience.}
\vspace{3pt}\\
\textsuperscript{1}DOTA 2 chat messages originally used to train a bot intended to mirror typical players' chat behavior\vspace{4pt}\\
\textsuperscript{2}DOTA 2 chat messages as well as other game data compiled by \href{https://www.opendota.com/}{Opendota}
}\label{datasets}
\end{table}

\subsection{DOTA 2}
\paragraph{Human Annotation}
An annotator labeled each message from 300 randomly sampled matches. When assessing a message, the annotator sees the previous messages that match, which player is the sender, and which team each player is on. A second annotator labeled a subsample of 20 matches; a PABA kappa of 0.7734 was measured between annotators, considered an expression of satisfactory agreement \citep{march2009pabak}. 

\subsection{\titleShort}
We selected a subset of chat messages from \titleFullName, consisting only of matches with at least 10 messages \rev{(see unlabeled \dsetcod{} in Table \ref{datasets})}. This data was only used for pretraining, not classification.

\paragraph{Human Annotation}
An annotator labeled each message from 200 matches occurring in a period of 1 week meeting the following conditions:
\begin{itemize}
    \item At least 10 messages per match
    \item Each player sent at least 3 messages per match
    \item Each player sent at least 50 messages total that week
\end{itemize}
For this dataset, the annotator sees the previous messages in a match, and which player sent each message. A second annotator labeled a subsample of 20 matches; a PABA kappa of 0.657 was measured between annotators, considered an expression of satisfactory agreement \citep{march2009pabak}.  Due to legal and privacy reasons, this data cannot be made public.

\subsection{Task and training}
We give our models previous messages in a conversation simply by concatenating them to the current message, although the downstream classifier's task is always to asses the toxicity of the current message. To use information on who sent each message, one can inject sender information explicitly by adding a new input, or by adding special tokens to the original input. \citet{yang2023toxbuster} find little difference in performance between these two approaches, we use the latter as it requires no modifications to the model architecture (see figure \ref{experiment_pipeline} part B). For consistency, the sender of the evaluated message is always labeled as player 0 from team 0. We explore domain-adaptive pretraining \citep{gururangan2020don} in the form of a masked language modeling (MLM) task (see figure \ref{experiment_pipeline} part A).

\subsection{Metrics and evaluation}
To address the label imbalance (toxicity is rare) we use a cost-sensitive loss function for all our classifiers, that gives a different weight to samples of each class \citep{fernandez2018cost}. In a downstream application, the decision threshold would also be adjusted based on an estimate of the cost of missing a toxic message versus mistakenly flagging a benign message as toxic \citep{handoyo2021varying}. We use weights proportional to the label distribution as this makes for straightforward comparisons with the balanced accuracy metric regardless of the underlying label distribution (chance level balanced accuracy is always 0.5). Following \cite{brodersen2010balanced} balanced accuracy is defined as:

\begin{equation}
\text{Balanced Accuracy} = \frac{1}{2} \left( \frac{TP}{TP + FN} + \frac{TN}{TN + FP} \right),
\end{equation}

where \( TP \) and \( FN \) denote true positives and false negatives, respectively, for the positive class, and \( TN \) and \( FP \) denote true negatives and false positives for the negative class. This formulation ensures that performance is evaluated independently of class imbalance. We also measure the Area Under the ROC Curve (AUC) as well as the Precision and Recall using scikit-learn implementation \footnote{https://scikit-learn.org/stable/api/sklearn.metrics.html}.

Our classifiers are each trained on a subset of 100 matches and tested on another 100 matches. For experiments involving pretraining, due to resource constraints, we stop at 10 epochs of MLM training \rev{following common practice as in \cite{bhardwaj2023pre}}. The downstream classifier metrics are averaged over 10 runs as the random order of the data and the model's initial state introduces variance, even though the dataset split remains unchanged across runs and models.

\subsection{Models}
\begin{figure}
\begin{center}
\includegraphics[width=\textwidth]{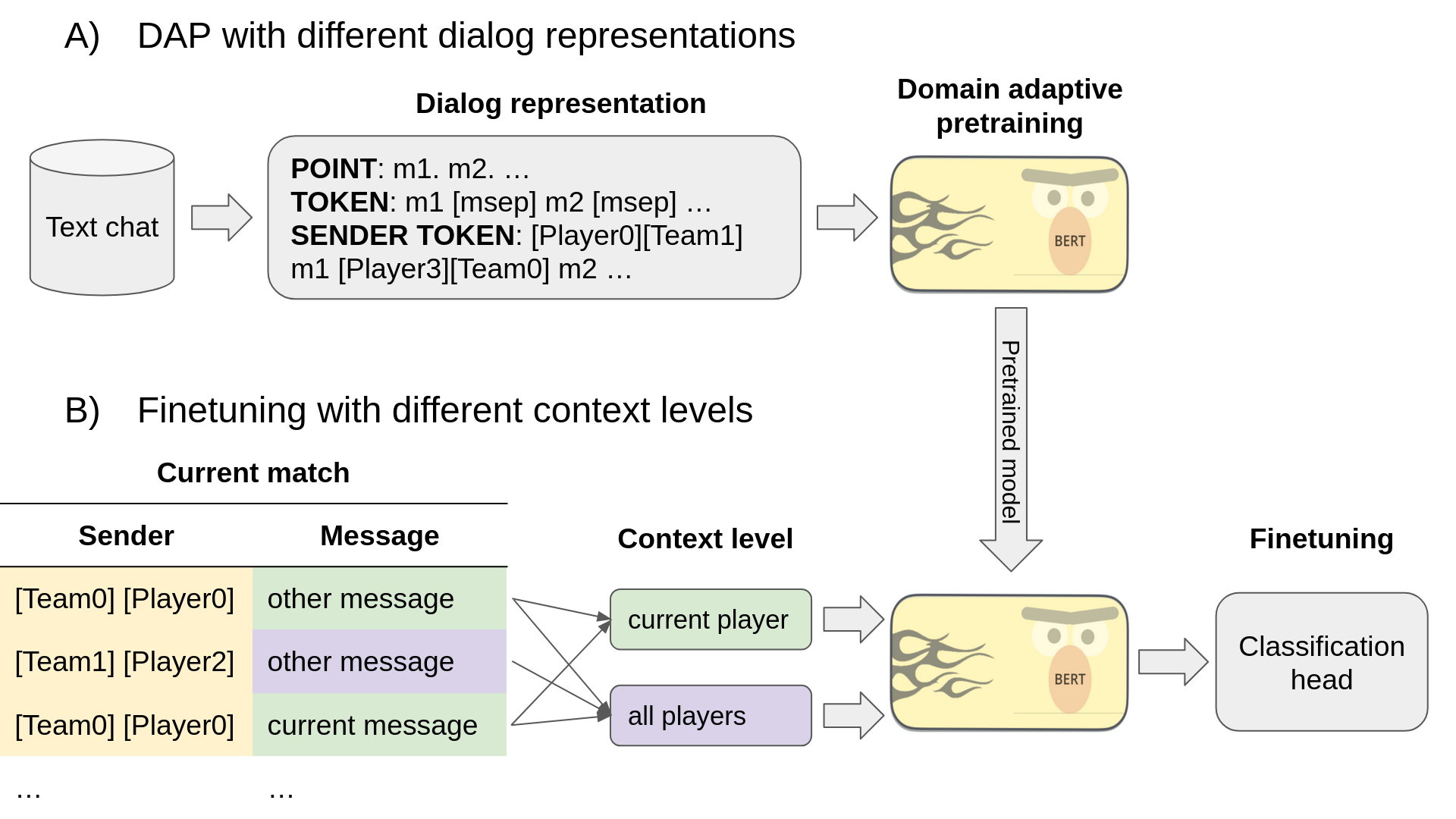}
\end{center}
\caption{
Experiment pipeline. In part A) we experiment with including metadata in domain adaptive pretraining. Part B) involves different context levels as shown in Figure \ref{context_levels}.
}\label{experiment_pipeline}
\end{figure}

We use distilRoBERTa-base \citep{sanh2019distilbert, liu2019roberta} as the base pretrained model in all experiments. We pretrain it further (figure \ref{experiment_pipeline} part A), and finetune it for classification according to HuggingFace’s standard sequence classification head (figure \ref{experiment_pipeline} part B).

\paragraph{Pretrained models:}
\begin{itemize}[leftmargin=*, itemsep=0.0mm, topsep=5pt]
    \item \textbf{\base{}}: distilRoBERTa-base was pretrained by \citet{sanh2019distilbert}, distilled from BERT. Weights downloaded from: \href{https://huggingface.co/distilbert/distilroberta-base}{https://huggingface.co/distilbert/distilroberta-base}.
    \item \textbf{\DAP{}}: Continued MLM pretraining of \base{} (domain-adaptive pretraining), messages separated by '.'
    \item \textbf{\sep{}}: Continued pretraining of \base{}, messages separated by special tokens added to the vocabulary that do not contain sender metadata.
    \item \textbf{\sender{}}: Continued pretraining of \base{}, with added special tokens for sender metadata (these also serve as separators)
\end{itemize}

\section{Results}


\begin{table}[t]
\centering
\resizebox{\textwidth}{!}{%
\large
\begin{tabular}{llclcl}
\toprule
Pretrained Model& Context&  Balanced Accuracy& Improvement& Precision& Recall\\
\midrule
\textbf{\dsetcod{}}& & & & & \\
\hline
 \base{}& & 0.72$_{\pm0.04}$& -& 0.30$_{\pm0.05}$& 0.58$_{\pm0.13}$\\

 \DAP{}& messages from& 0.70$_{\pm0.01}$& \redbox{\darr 0.02}& 0.30$_{\pm0.02}$& 0.52$_{\pm0.02}$\\

 \DAP{}\sep{}& all players& \textbf{0.73$_{\pm0.02}$} & \bluebox{\uarr 0.01}& 0.34$_{\pm0.04}$& 0.57$_{\pm0.01}$\\

 \DAP{}\sender{}& & 0.72$_{\pm0.01}$& -& 0.37$_{\pm0.02}$&0.52$_{\pm0.03}$\\
 \midrule
 
 \base{}& & 0.73$_{\pm0.04}$& -& 0.27$_{\pm0.06}$& 0.64$_{\pm0.14}$\\
 \DAP{}& messages from& 0.72$_{\pm0.02}$& \redbox{\darr 0.01}& 0.25$_{\pm0.01}$&0.62$_{\pm0.04}$\\
 
 \DAP{}\sep{}& current player& 0.73$_{\pm0.01}$& -& 0.31$_{\pm0.03}$&0.58$_{\pm0.01}$\\
 
 \DAP{}\sender{}& & \textbf{0.75$_{\pm0.01}$}& \bluebox{\uarr 0.02}& 0.38$_{\pm0.04}$&0.59$_{\pm0.02}$\\

\midrule

\textbf{\dsetdota}& & & & & \\
\hline
\base{}& & 0.65$_{\pm0.01}$& -& 0.11$_{\pm0.02}$& 0.53$_{\pm0.07}$\\

\DAP{}& messages from& 0.71$_{\pm0.02}$& \bluebox{\uarr 0.06}& 0.27$_{\pm0.03}$& 0.49$_{\pm0.04}$\\

\DAP{}\sep{}& all players& 0.66$_{\pm0.03}$& \bluebox{\uarr 0.01}& 0.13$_{\pm0.04}$& 0.49$_{\pm0.07}$\\

\DAP{}\sender{}& & \textbf{0.75$_{\pm0.02}$} & \bluebox{\uarr 0.10}& 0.23$_{\pm0.04}$& 0.59$_{\pm0.05}$\\

\midrule

 \base{}& & 0.71$_{\pm0.02}$& -& 0.15$_{\pm0.02}$& 0.58$_{\pm0.06}$\\

 \DAP{}& messages from& 0.73$_{\pm0.01}$& \bluebox{\uarr 0.02}& 0.18$_{\pm0.03}$& 0.59$_{\pm0.03}$\\

 \DAP{}\sep{}& current player& 0.75$_{\pm0.03}$& \bluebox{\uarr 0.04}& 0.19$_{\pm0.04}$&0.64$_{\pm0.06}$\\
 
 \DAP{}\sender{}& & \textbf{0.77$_{\pm0.02}$}& \bluebox{\uarr 0.06}& 0.17$_{\pm0.02}$&0.71$_{\pm0.05}$\\
 
\bottomrule
\end{tabular}
}
\caption{Metrics across datasets and experiments. Using metadata boosts balanced accuracy in 3 of 4 cases by 0.06 on average.}
\label{metrics}
\end{table}

\paragraph{Toxicity is highly contextual.} Half of the toxic messages in the \dsetdota{} dataset are context-independent, and only 11\% of \dsetcod{} toxicity is context-independent. It is worth noting that the \dsetcod{} dataset was conditionally sampled (matches with at least 10 messages) while the \dsetdota{} was uniformly sampled, which may impact label distributions.

\paragraph{The efficacy of Domain Adaptive Pretraining (DAP) is situational.} On average, experiments using DAP show better performance (Table \ref{metrics}) thanks to pretrained embeddings that are better suited for representing the slang, abbreviations, and misspellings common in game chat datasets. The \dsetcod{} dataset (upper half of Table \ref{metrics}) benefits much less from DAP. As mentioned previously, this can be largely attributed to the unobserved conversation missing from this dataset, but the lesser number of words and messages per match in this dataset may also play a role (see Table \ref{datasets}).

\paragraph{Metadata is useful in both pretraining and finetuning.} For both datasets, the best performing model uses metadata (see last row for each dataset in Table \ref{metrics}). Surprisingly, these best performing models have only messages from the current player as input, so the metadata tokens lose meaning. Preceding each message are special tokens that indicate who the sender is, but if all messages are from the current player, the sender is always '[Team0][Player0]' and these special tokens seem to serve no other purpose than separating messages. However, experiments using separator special tokens that do not contain metadata (rows with \DAP{}\sep{} in Table \ref{metrics}) show worse performance. The difference lies in the pretrained embeddings: for domain adaptive pretraining, messages from all players are always present, and including metadata here results in better downstream performance. We conclude that using metadata in pretraining results in higher quality embeddings. This result informs the pretraining/finetuning paradigm: pretraining with additional information specific to the downstream task can boost performance. This is especially useful when few labeled samples are available for finetuning, as is the case in this work.

\paragraph{More context is not always better.} Experiments using only the current player's messages show consistently better performance than experiments using messages from all players (see Table \ref{metrics}). More context leads to worse performance in this case, likely because there are too few labeled samples to learn the relationships between messages from different players, and correctly attribute toxicity to a specific player. Adding messages from other players can be seen as multiplying the number of input features. However, the number of training samples does not increase proportionally, resulting in far more features than training samples.

\paragraph{Long-term player behavior is predictive of current toxicity.} The labeled \dsetcod{} dataset was sampled from a 1-week period, and the players present in this dataset also appear in other matches within this same period that were not labeled. \rev{As such our dataset is a subset of the data and does not represent the entirety of the player volume or experience.} By applying the best classifier to these unlabeled matches, we create a toxicity propensity score for each player. This propensity score is the average predicted label of all messages sent by the player in the 1-week period, excluding those already manually labeled. By fitting a linear classifier (Linear regression, SVM, perceptron) on this propensity score, we obtain up to 74\% balanced accuracy in predicting the toxicity of test set messages. In other words, by labeling all messages from toxic players as toxic, and all messages from non-toxic players as not toxic, one can achieve classification performance far above the chance level. Note that the model responsible for the propensity score was not trained on any messages written by \textbf{any player present in the test dataset}.

\section{Limitations}
While we conducted multi-annotator agreement estimation, we did not perform additional steps to align final labels with human judgments through iterative refinement of the toxicity definition. While our labeled data provides a useful basis for comparing models and methods, it is not intended for direct deployment in a live system. Furthermore, the data contains many different languages, and samples not in the annotator’s native languages (English and French) are expected to be of lower quality. Our primary goal was not to push the state-of-the-art in toxicity detection but rather to explore different methods for incorporating context and adapting to game-specific speech. Given these objectives, we focused on efficient experimentation using small, monolingual models, a limited amount of labeled data, and a small number of training iterations.

\section{Discussion}
We have observed the importance of context regarding ground truth annotation, in that a large portion of toxicity was labeled as such only because of context. In other cases, the context is entirely unneeded by the annotator and plays no role in their decision. Although some work in XAI has explored the differences between annotator and AI classification \citep{venkatesh2024comparing}, it remains unclear if context plays an analogous role in the decisions of machine learning models, and to what extent AI can capture highly context-dependent samples.

Highly context-dependent toxicity was much more common in the \dsetcod{} dataset, and off-the-shelf solutions not designed to use context do not usually perform well here. It is possible that toxic players produce more context-dependent toxicity precisely to evade moderation from such off-the-shelf solutions: Activision employs one to filter chat proactively, while \dsetdota{} chat in our dataset is entirely unfiltered and has less context-dependent toxicity.

The viability of using players' long term behavior to predict toxicity suggests that some players are more prone to being toxic (at least in a given time period). If these people can be acted on (e.g. encouraged to be more prosocial) platforms could potentially alleviate overall toxicity issues without needing direct moderation efforts such as censoring \citep{wijkstra2023help}.

The more context is used for toxicity detection, the greater the need for information to be monitored or recorded. This is expensive \citep{li2020qualitative} and not conducive to users' privacy, as such it is important to know which sources of context are most useful for toxicity detection, and which can be discarded.

\section{Future work}
To perfectly assess the toxicity of an interaction, the moderator must witness the interaction firsthand and have access to the exact information the participants have. While this ideal scenario is unrealistic, aside from text and voice chat there is \rev{an} abundance of contextual information unused in both classification and annotation. In fact, verbal communication is not even required to be toxic, as in-game actions can be toxic in and of themselves \citep{canossa2021honor}. 

There is potential to use long term player behavior to boost performance by representing it pertinently, and incorporating it into training. We attempted to train a classification head that receives frozen embeddings of past matches in addition to the current match. Although no improvement was achieved compared to using only the current match, other approaches (especially using more labeled data) seem promising. Furthermore, a user's long term behavior can be modeled by more than their posts or messages, users often have profiles with statistics and social networks.

\section{Conclusion}
In this work, we present a method to employ existing pretrained models for context-sensitive toxicity detection without changing its architecture. We use contextual information on past messages and their senders, incorporating the latter in domain-adaptive pretraining, and show the viability of using players' long term behavior for toxicity detection. Our work puts much more emphasis on proactive moderation (acting as a message is posted) compared to previous studies that operate largely in hindsight, and we present new sources of context with new ways of using them. Overall we highlight the importance of context, especially for ground truth labels, where annotators' access to context has often been unreported, inconsistent, or unclear. We provide the code used to run these experiments at \href{https://github.com/AdrienSF/proactive-contextful-toxicity-detection}{https://github.com/AdrienSF/proactive-contextful-toxicity-detection}.

\section*{Acknowledgments}
TBA

\section*{Ethics Statement}
This study focuses on the detection of toxicity in online multiplayer games, a subject that inherently involves sensitive issues related to content moderation, privacy, and fairness in automated decision-making. All data used in this research was either publicly available or obtained with appropriate permissions, ensuring compliance with ethical and legal guidelines for data handling. Given the potential societal impact of automated moderation systems, we acknowledge the risks of over-moderation, misclassification, and biases in toxicity detection. To mitigate these risks, we emphasize the importance of context-aware models and transparent methodologies. Additionally, our findings underscore the necessity for human oversight in automated moderation processes, particularly in environments with nuanced and domain-specific language. We recognize that defining toxicity is inherently subjective, and efforts were made to adopt a well-established framework while acknowledging its limitations. Future research should explore how these models impact different communities and develop strategies for inclusive, fair, and interpretable moderation systems.

\bibliography{colm2025_conference}
\bibliographystyle{colm2025_conference}

\appendix
\section{Appendix}

\begin{table}[h]
\begin{center}
\begin{tabular}{ p{.2\linewidth} p{.7\linewidth} }
    \toprule
    \textbf{Toxicity Level} & \textbf{Description of level} \\
    \midrule
    Very Toxic & A comment that is very hateful, aggressive, disrespectful, or otherwise very likely to make a user leave a discussion or give up on sharing their perspective. \\
    \midrule
    Toxic & A comment that is rude, disrespectful, unreasonable, or otherwise somewhat likely to make a user leave a discussion or give up on sharing their perspective. \\
    \midrule
    Not Toxic & A neutral, civil, or even nice comment very unlikely to discourage the conversation. \\
    \midrule
    I'm not sure & The comment could be interpreted as toxic depending on the context but you are not sure. \\
    \bottomrule
\end{tabular}
\end{center}
\caption{Google's perspective API definition of toxicity}\label{perp_tox_def}
\end{table}

\begin{table}[h]
\begin{center}
\resizebox{\textwidth}{!}{
\begin{tabular}{llllllllllll}
\toprule
   base model & context & \multicolumn{2}{c}{balanced accuracy }& \multicolumn{2}{c}{AUC}& \multicolumn{2}{c}{binary F1}& \multicolumn{2}{c}{precision}& \multicolumn{2}{c}{recall}\\
    &  & mean & std & mean & std & mean & std & mean & std & mean & std \\
 \midrule
 \textbf{\dsetcod{}} & & & & & & & & & & & \\
 \midrule
 
  \base{} &  & 0.72 & 0.04 & 0.81 & 0.02 & 0.39 & 0.04 & 0.30 & 0.05 & 0.58 & 0.13 \\
 
  \DAP{} & messages from & 0.70 & 0.01 & 0.77 & 0.03 & 0.38 & 0.02 & 0.30 & 0.02 & 0.52 & 0.02 \\
  
  \DAP{}\sep{} & all players & \textbf{0.73} & 0.02 & 0.79 & 0.03 & 0.43 & 0.04 & 0.34 & 0.04 & 0.57 & 0.01 \\
  
  \DAP{}\sender{} &  & 0.72 & 0.01 & \textbf{0.81} & 0.01 & 0.43 & 0.02 & 0.37 & 0.02 & 0.52 & 0.03 \\
  
\midrule
  
  \base{} &  & 0.73 & 0.04 & 0.81 & 0.02 & 0.37 & 0.05 & 0.27 & 0.06 & 0.64 & 0.14 \\
  
  \DAP{} & messages from & 0.72 & 0.02 & 0.82 & 0.02 & 0.36 & 0.02 & 0.25 & 0.01 & 0.62 & 0.04 \\
  
  \DAP{}\sep{} & current player & 0.73 & 0.01 & 0.82 & 0.01 & 0.40 & 0.03 & 0.31 & 0.03 & 0.58 & 0.01 \\
  
  \DAP{}\sender{} &  & \textbf{0.75} & 0.01 & \textbf{0.85} & 0.01 & 0.46 & 0.03 & 0.38 & 0.04 & 0.59 & 0.02 \\
  
\midrule

 \textbf{\dsetdota{}} & & & & & & & & & & & \\
 \midrule
  
  \base{} &  & 0.65 & 0.01 & 0.71 & 0.02 & 0.17 & 0.02 & 0.11 & 0.02 & 0.53 & 0.07 \\
  
  \DAP{} & messages from & 0.71 & 0.02 & 0.86 & 0.02 & 0.35 & 0.03 & 0.27 & 0.03 & 0.49 & 0.04 \\
  
  \DAP{}\sep{} & all players & 0.66 & 0.03 & 0.76 & 0.05 & 0.20 & 0.04 & 0.13 & 0.04 & 0.49 & 0.07 \\
  
  \DAP{}\sender{} &  & \textbf{0.75} & 0.02 & \textbf{0.86} & 0.02 & 0.33 & 0.04 & 0.23 & 0.04 & 0.59 & 0.05 \\

\midrule

  \base{} &  & 0.71 & 0.02 & 0.79 & 0.01 & 0.23 & 0.02 & 0.15 & 0.02 & 0.58 & 0.06 \\
  
  \DAP{} & messages from & 0.73 & 0.01 & 0.85 & 0.03 & 0.27 & 0.03 & 0.18 & 0.03 & 0.59 & 0.03 \\
  
  \DAP{}\sep{} & current player & 0.75 & 0.03 & 0.85 & 0.04 & 0.29 & 0.04 & 0.19 & 0.04 & 0.64 & 0.06 \\
  
  \DAP{}\sender{} &  & \textbf{0.77} & 0.02 & \textbf{0.87} & 0.01 & 0.27 & 0.02 & 0.17 & 0.02 & 0.71 & 0.05 \\

\bottomrule
\end{tabular}
}
\end{center}
\caption{All metrics.}\label{full_metrics}
\end{table}

\begin{figure}[h]
\begin{center}
\includegraphics[width=.75\textwidth]{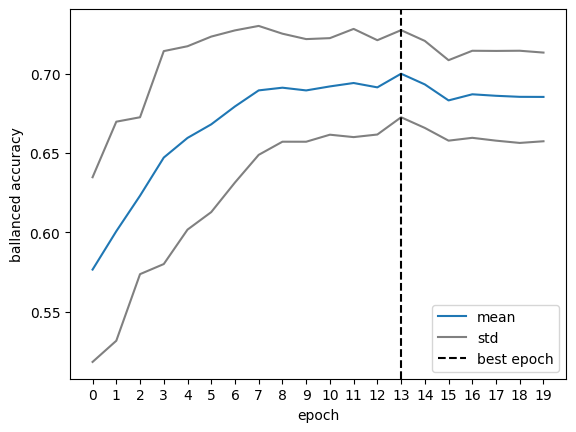}
\end{center}
\caption[]{Example of epoch selection for the experiment using \base{} and messages from current player with learning rate 5e-6. Metrics from all other experiments can be viewed in the same format at \href{https://github.com/AdrienSF/proactive-contextful-toxicity-detection}{https://github.com/AdrienSF/proactive-contextful-toxicity-detection}.
}
\label{example_epoch_selection}
\end{figure}

\paragraph{Hyperparameter tuning.} For each experiment, after 10 training runs, we present all metrics measured just after the epoch that shows the best mean balanced accuracy (see figure \ref{example_epoch_selection}). We try several initial learning rates for each experiment and present metrics for the learning rate that gives the best maximum mean balanced accuracy across runs and epochs. A cosine learning rate scheduler was used with a period 2 times the number of total epochs. Metrics for all epochs and learning rates can be viewed at \href{https://github.com/AdrienSF/proactive-contextful-toxicity-detection}{https://github.com/AdrienSF/proactive-contextful-toxicity-detection}.

\end{document}